\documentclass{article}
\usepackage{spconf,amsmath,graphicx}
\usepackage{hyperref}       

\newcommand{\orcid}[2]{\href{https://orcid.org/#1}{#2\hspace{0.5mm}\includegraphics[scale=0.06]{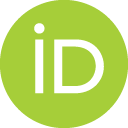}}}

\title{A Review on Oracle Issues \\
in Machine Learning}
\name{\orcid{0000-0003-2652-3718}{Diogo Seca}\thanks{This work is financed by National Funds through the Portuguese funding agency, FCT - Fundação para a Ciência e a Tecnologia, within project UIDB/50014/2020.}}
\address{LIAAD - INESC TEC\\
Porto, Portugal}

\begin{document}
\maketitle

\begin{abstract}
    Machine learning contrasts with traditional software development in that the oracle is the data, and the data is not always a correct representation of the problem that machine learning tries to model. 
    We present a survey of the oracle issues found in machine learning and state-of-the-art solutions for dealing with these issues. These include lines of research for differential testing, metamorphic testing, and test coverage. We also review some recent improvements to robustness during modeling that reduce the impact of oracle issues, as well as tools and frameworks for assisting in testing and discovering issues specific to the dataset.
\end{abstract}


\section{Introduction}
    The topic of testing machine learning systems has recently gained significant attention in the literature \cite{Riccio2020-sn}.  
    
    In traditional software development, oracle tests are implemented as tests that assert whether a given input produces an expected output \cite{Barr2015-xa}. In machine learning (ML), a learning algorithm might be bug-free but still deviate from the desired behavior due to faults during training. Machine learning contrasts with traditional software development in the sense that the data is considered the oracle. Moreover, a machine learning model heavily depends on data \cite{Amershi2019-dl}. Bugs in data affect the quality of the generated model and can be amplified to yield more serious problems over a period of time \cite{breck2019-dv}. 
    

    \subsection{Data in Machine Learning}
    For supervised learning problems, the data is usually segregated into two groups:
    
    (1) \textbf{Training data}. A portion of the data is used to learn a model according to a particular learning algorithm such as XGBoost \cite{Chen2016-oq};
    
    (2) \textbf{Test data}. The remaining portion of the data is used to evaluate the predictions of the learned model using metrics. In classification problems, our predicted variable (class) is categorical (or binary, in the special case of binary classification), which is often evaluated using metrics such as Accuracy, Precision, Recall, F1, or AUC. In regression problems, our predicted variable (target) is continuous, which is often evaluated using metrics such as Mean Absolute Error, Mean Squared Error, or R-Squared.
    
    The occurrence of missing values is common in real datasets in supervised machine learning. However, there is a branch of machine learning, named semi-supervised learning \cite{Van_Engelen2020-bw}, which aims to deal with problems in which only a portion of the data (oftentimes small) contains information regarding the target.
    In unsupervised learning, there is no target, which means there is no particular variable that we want to predict. Instead, we seek to learn a model from the patterns in the data that correctly represents the probability densities over inputs. One recent example of unsupervised learning is generative models such as Generative Adversarial Networks \cite{Goodfellow2014-rz}.

    \subsection{The novelty of Machine Learning Testing}
    
    Most of the research in machine learning testing is centered on supervised learning, some research is focused on issues such as correctness and robustness \cite{Zhang2020-ns}. Few papers relate to testing interpretability, privacy, or efficiency \cite{Zhang2020-ns}.

    Researchers have recently started adapting concepts from the software testing domain to help the machine learning community detect and correct faults in ML programs, namely code coverage, mutation testing, and metamorphic testing \cite{Braiek2020-wf} \cite{Zhang2020-ns}. In order to improve machine learning software, Amershi et al. suggest not only focusing on traditional software debugging but also on error analysis \cite{Amershi2019-dl}.
    
    \subsection{The Oracle Problem}
    
    The oracle problem refers to situations where it is extremely difficult, or impossible, to verify the test result of a given test case (that is, an input selected to test the program) \cite{Chen2018-oo}.
    
    \cite{Groce2014-xx} argue that the only possible oracle is the end-user, and so interactive ML systems are important to converge towards the oracle truth. However, the oracle is known beforehand in the case of synthetic time series, except for the randomness introduced.
    
    The identification of test oracles remains challenging because many desired properties are difficult to formally specify and domain-specific knowledge is often required. Companies usually rely on third-party data labeling companies to get manual labels, which can be expensive. Alternatively, metamorphic relations are a type of pseudo oracle adopted to automatically mitigate the oracle problem in machine learning testing \cite{Zhang2020-ns}.
    
    \subsection{Contributions}
    
    Our research follows the previous work of \cite{Liem2020-cq}. Two gaps existed in the literature which we hope to contribute towards, namely a survey of the oracle issues found in machine learning and the state of the art solutions for dealing with these issues. Section 2 delves into the common issues found in machine learning, and briefly discusses some solutions. Sections 3, 4, and 5 present three strategies and best practices that solve most of these issues. Section 6 concludes this work with a foreword for future research on this topic.

\section{Common Issues}

    

    \subsection{Relating to data}
    
    The following are the common issues in the data which, despite not being caused by the machine learning practitioner, may still negatively impact the modeling process. These are issues that the machine learning practitioner cannot control, but its impact may be minimized by sticking to reasonable practices. 
    
    \vspace{1em}
	\textbf{Noise in data}
	
	Noise can occur in label/target, as well as noise in the features. Noise in label/target can also be conditional on the features, and be the result of an unknown exogenous variable at play. If this exogenous variable is known, then the ideal solution would be to gather data on it and find a way to incorporate this data within the modeling process. However, this is not always the case. As such, generally, machine learning pipelines should be resilient to noise found in data.

    \vspace{1em}
    \textbf{Partially missing data}
    
    The data may have missing values in some features. This is automatically imputed by some machine learning pipelines, while in others, a new binary feature is created to introduce the information that there is a missing value in a given feature. 
    Another form in which crucial data might be missing is if an entire feature is missing. Imagine predicting temperatures without knowing whether it was summer or winter, or even what the day of the year was. 

    \vspace{1em}
	\textbf{Ambiguous data}
	Another is where the data is ambiguously annotated, fooling practitioners into making irrelevant assumptions.

    \vspace{1em}
	\textbf{Unrepresentative samples}
	
	The dataset may not be a good representation of the real world. For example, sampling electors who only use landline may result in a biased sample. By feeding this biased sample to a learning algorithm, we may be learning a biased model. A common sub-issue is unbalanced datasets, in which the label/target is severely skewed towards some particular values. Some sampling techniques like SMOTE may help reduce this effect, resulting in better predictions.

    \vspace{1em}
	\textbf{Irrelevant raw data}
	
	Some methods such as k-Nearest Neighbors scale poorly when increasing the number of features. This is because k-NN is a distance-based algorithm, and when the number of features increases, the significance of distant values in some dimensions starts to decrease.
    
    \subsection{Relating to the use of data}

	\vspace{1em}
	\textbf{Poor framing of the problem}

	Some problems are difficult to frame. 
	An example of such cases is attempting to model trends in time series data using k-Nearest Neighbors. When given data with a simple monotonic trend, k-Nearest Neighbors looks at the local average, which is an insufficient assumption to generalize the trend component in time series. A solution to this issue would be to instead of forecasting the time series in raw, forecast the time series slope. Transformations such as these constitute a change in the distribution of the variable(s) we wish to model and predict.
	Another common problem is when outliers are present, and the modeling process assumes distributions such as the normal distribution that is unable to account for outliers. As such, outliers end up having a significant impact on the modeling process, resulting in poor generalization. 

	\vspace{1em}
	\textbf{Poor modeling assumptions}
		
	Assumptions can be made to improve the modeling process. However, this results in learning a model significantly different. However, these assumptions add information where the data may not. For example, Convolution Neural Networks (CNNs) assume that vanilla networks do not: they assume that some patterns are equally important, no matter where they show in the picture. This changes the context of optimization and will result in a differently learned neural network model.
	In computer vision, Generative Adversarial Networks have become known for overcoming correcting this issue where Variational Auto-Encoders could not \cite{Goodfellow2014-rz}. Generative Adversarial Networks do so by adapting their objective function dynamically, enabling learning complex distributions present in the data \cite{Goodfellow2014-ss}.

	\vspace{1em}
	\textbf{Misinterpretations of the data}

	It's a common mistake to eyeball correlations and patterns when looking at two pairs of time series. Humans are naturally gifted at finding patterns, even when there are none, and the time series are resultant from a white noise generator.

	\vspace{1em}
	\textbf{Irrelevant features}

	Data can be turned into features, but not all features are equally important. Features can be quantified in terms of feature importance using permutation testing or Shapley values.

	If the number of features is too large, an optimized selection of features can be efficiently found using Recursive Feature Elimination.

	Another approach that may help tackle the problem is applying dimensionality reduction techniques such as Principle Component Analysis \cite{wold1987principal} or Auto-Encoders \cite{vincent2008extracting}.

	\vspace{1em}
	\textbf{Multiple hypothesis testing}

	Optimization is often used to squeeze the best out-of-sample performance when using a learning algorithm that requires the input of hyperparameters. However, when the search hyper-parameter increases, so to does the likelihood of finding the perfect fit for our validation set (the set we validate the choice of hyper-parameters). A reasonable solution for such cases is to leave a portion of unseen data for one final test out-of-sample.
	
	Another way that might help reduce this issue is using Stratified Cross-Validation, and Bayesian Optimization (e.g. Optuna \cite{optuna_2019}) to sample the hyper-parameter solution space conservatively.
	When testing for statistical significance, the Bonferroni correction may be applied in order to adjust the significance threshold to the number of tests performed \cite{Dunn61multiplecomparisons}.

\section{Differential Testing}

    Differential testing is a software testing technique to detect bugs in the implementation by providing the same input to a series of similar applications or different implementations of the same application, and observing differences in their execution \cite{mckeeman1998differential}. Differential testing can be applied to machine learning by either using different learning algorithms or by using different backend implementations of the same algorithm. 
    
    Different learned models are said to be experts at specific subsets of the data. Liem and Panichella learned different models using distinct learning algorithms, and then calculate the entropy between their predictions for the same instances \cite{Liem2020-cq}. Entropy is higher for instances where the predictions of the different models don't agree on. This heuristic allowed them to identify the potential oracle problems in deep learning system, such as finding ambiguous labels and data hard to model (due to observability issues). 
    
    \subsection{Ambiguous data labeling}
    
    In computer vision, when multiple models consistently have problems recognizing the ground truth, the image class may not visually stand out. One example is during the classification of velvet, which doesn’t stand out as it is a material, rather than a truly recognizable object \cite{Liem2020-cq}. 

    When two classes are consistently confused by the models they are synonyms, homonyms, or meronyms. One such solution learning from data with synonym labels is to group them under a single label. For homonyms, it is to change the label in order to distinguish distinct terms. Meronyms may occur when an image that contains one object/label might also contain objects relating to the other paired label. Meronyms can be analyzed in hierarchies, and if required apply hierarchical methods \cite{Liem2020-cq}.

    If we're training a model to detect the most salient object in an image, what should the model predict, from an image showing a bucket with oranges? A bucket? Oranges? Labeling these images using only one of the terms is subjective and may lead to ambiguity. If for a given image class, multiple models consistently have problems recognizing the ground truth class, the image class may not visually stand out \cite{Liem2020-cq}.

    \subsection{Class independence}

    Oftentimes, in image classification problems, the classes are dependent \cite{Liem2020-cq}. However, they are mathematically represented as independent. Moreover, the maximum likelihood criteria makes no assumptions of dependence between variables. A chihuahua could be considered equally far away from a bicycle as a bulldog. 

    During the training of an ML classification pipeline, the common criterion to optimize for is the likelihood of the ground truth class, which should be maximized. With a single ground-truth label being available per image, the best result in terms of optimization, therefore, is to have a prediction confidence of 1.0 for a single class (and thus, a probability of 0.0 for other classes), even if multiple classes are present. Thus, while a beach wagon typically contains more than one car wheel, if the first class was the ground truth, optimization is considered to have succeeded better if an ML system classifies beach wagon with 1.0 confidence, thus being ‘blind’ to the possible presence of car wheels \cite{Liem2020-cq}.

    This could be fixed by taking prediction confidence into consideration. Probabilistic machine learning, which we use to predict probability distributions with uncertainty could be a solution. Probabilistic learning also avoids the maximum likelihood criteria as it does not use the maximum likelihood estimation as the objective function. Instead, it uses maximum a posteriori probability. Probabilistic models can be beneficial including calibration and when dealing with missing data in medical applications \cite{chen2020probabilistic}.

    \subsection{CRADLE}

    CRADLE is a tool for testing Deep learning applications, inspired by differential testing \cite{Pham2019-yq}. CRADDLE uses different Deep Learning backends and compares predictions of their learned models. Some discrepancies point to implementation faults in the backend. However, the existence of discrepancies may signify that a label is inconsistent, so human supervision is still required to distinguish between these two scenarios. The authors also point to the generation of mutated models using fuzzing to generate more tests.

    \subsection{Audee}
    
    Audee, is another differential testing approach based on CRADLE that has been used to discover 26 unknown bugs in Deep Learning frameworks \cite{Guo2020-ih}. Audee adopts a search-based approach and implements three different mutation strategies to generate diverse test cases by exploring combinations of model structures, parameters, weights and inputs. Audee can detect three types of bugs: logical bugs, crashes and Not-a-Number (NaN) errors.

\section{Metamorphic Testing}


    Xie et al. point out that cross-validation alone is not sufficiently effective to detect faults in a supervised classification program. These authors propose using metamorphic testing as an additional validation technique in order to alleviate the oracle problem \cite{Xie2011-vx}.
    
    Metamorphic testing is a property-based software testing technique, which can be an effective approach for addressing the test oracle problem and test case generation problem \cite{Segura2016-fq}. For the specific case of machine learning, metamorphic relationships change the training data in some specific way and then analyze the changes in the output of the retrained system. 
    
    Data augmentation techniques like scaling, cropping, and rotating are commonly used to augment datasets for image analysis \cite{shorten2019survey}. Nair et al. claim to be the first to explore the area of data augmentation techniques for ML-based analysis of software code \cite{Nair2019-lt} . Using data with diversity to train models leads to more robust models \cite{Nakajima2018-qc} \cite{Nakajima2019-wa}.

    \subsection{Metamorphic relations}

    Metamorphic relations are necessary properties of the target function or algorithm in relation to multiple inputs and their expected outputs \cite{Chen2018-oo}.
    
    When testing big data software, metamorphic relations not only cover properties of the system under test but may also cover properties of the data itself. Similar to the program-related properties, these data-related properties can help produce additional follow-up data to form the sample data, and to verify the test results, especially when the oracle problem exists \cite{Alexandrov2013-mn}.
    
    The first step is to identify a set of properties or metamorphic relations that relate to multiple inputs and their outputs of the algorithm for the target program. The source test cases and their corresponding follow-up test cases are constructed based on these metamorphic relations. Following the executing of all these test cases using the machine learning pipeline, we can check whether the outputs satisfy their corresponding metamorphic relations \cite{Xie2011-vx}.
    
    Based on the literature, we have 21 listed metamorphic relations for machine learning systems \cite{Murphy2008-ai}\cite{Murphy2010-pa}\cite{Xie2011-vx}\cite{Nakajima2019-cj}\cite{Herbold2020-xf}:
    
    \begin{enumerate}
        \item Permutation of the order of elements in a set;
        \item Taking the inverse of each element in a set;
        \item Consistence with affine transformation $f(x) = kx + b, (k \neq 0)$;
        \item Permutation of class labels;
        \item Permutation of the attribute;
        \item Addition of uninformative attributes (e.g. a constant attribute);
        \item Addition of informative attributes;
        \item Multiplying numerical values e.g. by a constant;
        \item Consistence with re-prediction;
        \item Addition of training samples;
        \item Addition of classes by duplicating samples;
        \item Addition of classes by re-labeling samples;
        \item Removal of classes;
        \item Removal of samples;
        \item Reduce Margin (e.g. cropping an image);
        \item Insert noise;
        \item Insert separable;
        \item Insert inputs with extremely large values;
        \item Insert with all features exactly zero;
        \item In case of classification, insert empty classes;
        \item If categorical data is supported, insert inputs with empty categories in categorical data.
    \end{enumerate}
    
    These relations led to the discovery of bugs in the machine learning tools used Properties of machine learning applications for use in metamorphic testing \cite{Murphy2008-ai}. 
    
    \cite{Xie2011-vx} and \cite{Xie2009-zb} argue that metamorphic relations may represent both necessary and expected properties of the algorithm under test: (1) violations of necessary properties are caused by faults in the algorithm and therefore are helpful for the purpose of verification; (2) violations of expected properties indicate divergences between what the algorithm does and what the user expects, and thus are helpful for the purpose of validation.
    
    \cite{Nakajima2016-vh} designed a method that extracts metamorphic properties using support vector machines. The authors distinguish between metamorphic relations for creating pseudo oracles and metamorphic relations for data generation. However, the distinctions between these concepts seem redundant as they both end up generating more data and more testing cases.
    
    \subsection{Tools}

    \textbf{Dataset coverage.} 
    Zhang et al. studied test adequacy criteria used to provide quantitative measurement on the degree of the target software that has been tested, e.g.: line coverage, branch coverage, dataflow coverage \cite{Zhang2020-ns}. Nakajima and Ngoc Bui proposed a new coverage metric called \emph{dataset coverage} for the testing of machine learning programs \cite{Nakajima2016-vh}. Dataset coverage focuses on the characteristics of the population distribution in the training dataset. This metric is generated by testing newly generated data points based on metamorphic relations. However, this metric has some drawbacks: 
    \begin{enumerate}
        \item complete dataset coverage is not possible because an infinite number of data points can exist in between any pair of adjacent data points.
        \item the number of possible populations in datasets is also infinite. 
        \item it requires parameterization.
    \end{enumerate}

    \textbf{Amsterdam framework.} 
    Murphy et al. presented a framework named Amsterdam for the automated application of metamorphic testing \cite{Murphy2010-pa}. The tool takes as inputs the program under test and a set of metamorphic relations, defined in an XML file. 
    
    The authors argue that in certain cases slight variations in the outputs are not actually indicative of errors, e.g., floating-point calculations. To address this issue, Murphy et al. propose the concept of heuristic test oracles, by defining a function that determines whether the outputs are “close enough” to be considered equals \cite{Murphy2010-pa}.
    

\section{Robust Modeling}

    
    
    Currently, metamorphic testing is already being developed into some of the most recently deep learning algorithms like Generative Adversarial Networks \cite{Goodfellow2014-rz}, who have resulted in state-of-the-art results in computer vision \cite{Yang20173DOR} and reinforcement learning \cite{Henderson2018OptionGANLJ}. Generative Neural Networks are trained after several iterations of adversarial attacks, in order to minimize the error of the generated samples. This includes the application of transformations of the input space, e.g. noise and rotations.

    Humbatova et al. report that the most frequently reported issues related to the training of deep learning models are the quality of the data, and preprocessing \cite{Humbatova2020-pe} This section is dedicated to the improvements that can be made to the machine learning pipeline a priori in order to make the pipeline more robust to oracle issues. We consider the machine learning pipeline to include data preprocessing, training, and prediction.
    
    The No free lunch theorem suggests that there is no ultimate machine learning algorithm and that each model works best at some datasets \cite{Whitley1970-zu}. However, the latest improvements in machine learning have been mostly due to the same learning algorithms, and some even propose that a general-purpose algorithm is indeed possible by the application of meta-learning \cite{Giraud-Carrier2005-hi}.
    
    
    With the goal of minimizing oracle issues, Herbold et al. suggest using grid search in order to achieve equivalence class coverage for the hyperparameters \cite{Herbold2020-xf}. Instead, we suggest using bayesian optimization instead as it is more effective at navigating the hyperparameter search space based on bayesian theory \cite{snoek2012practical}.

    \subsection{Monitoring during modeling}
    
    When a new model shows better aggregate performance, it can be hard to notice whether its performance worsened on specific types of inputs. Testing predictions for a specific hand-picked set of inputs is also crucial as this helps guarantee that any model we use always produces the expected output on these example inputs \cite{Ameisen2020-sm}.
    
    \subsection{Monitoring after deployment}
    
    In order to guarantee the performance after modeling, \cite{Breck2017-ne} present 8 indicators for continuous monitoring after deployment of the machine learning pipeline. 


\section{Conclusion}


    We have tackled several research paths that have tried to solve or at least minimize the impact of oracle issues.
    
    Given the recent exposure of oracle issues and the advance of testing practices for machine learning, we anticipate the creation of novel techniques robust to some of the most common issues. We pose the question of how can meta-learning use these prior experiences and knowledge to improve future modeling.
    

\bibliographystyle{IEEEbib}
\bibliography{refs}
\end{document}